\documentclass[letterpaper]{article} 
\usepackage{aaai24}  
\usepackage{times}  
\usepackage{helvet}  
\usepackage{courier}  
\usepackage[hyphens]{url}  
\usepackage{graphicx} 
\usepackage{natbib}  
\usepackage{caption} 
\usepackage{algorithm}
\usepackage{diagbox}
\usepackage{makecell}
\usepackage{amssymb}
\usepackage{amsmath}
\usepackage{algpseudocode}
\usepackage{array}
\usepackage{color}
\usepackage{booktabs}

\usepackage[colorlinks,
            linkcolor=red, 
            citecolor=green, 
            ]{hyperref}

\usepackage{newfloat}
\usepackage{listings}
\DeclareCaptionStyle{ruled}{labelfont=normalfont,labelsep=colon,strut=off} 
\floatstyle{ruled}
\newfloat{listing}{tb}{lst}{}
\floatname{listing}{Listing}
\title{${\rm CA}^{2}$: Class-Agnostic Adaptive Feature Adaptation for One-class Classification}
\author{
    Zilong~Zhang\textsuperscript{\rm 1},
    Zhibin~Zhao\textsuperscript{\rm 1},
    Deyu Meng\textsuperscript{\rm 2},
    Xingwu Zhang\textsuperscript{\rm 1}\thanks{Corresponding author},
    Xuefeng Chen,\textsuperscript{\rm 1}
}
\affiliations {
    \textsuperscript{\rm 1} School of Mechanical Engineering, Xi’an Jiaotong University\\
    \textsuperscript{\rm 2} School of Mathematics and Statistics, Xi’an Jiaotong University\\
    zhangzilongc@gmail.com; zhaozhibin@xjtu.edu.cn; dymeng@mail.xjtu.edu.cn; xwzhang@mail.xjtu.edu.cn; chenxf@xjtu.edu.cn
}

\usepackage{bibentry}

\begin{document}

\maketitle

\begin{abstract}
One-class classification (OCC), i.e., identifying whether an example belongs to the same distribution as the training data, is essential for deploying machine learning models in the real world. Adapting the pre-trained features on the target dataset has proven to be a promising paradigm for improving OCC performance. Existing methods are constrained by assumptions about the number of classes. This contradicts the real scenario where the number of classes is unknown. In this work, we propose a simple \textbf{c}lass-\textbf{a}gnostic \textbf{a}daptive feature adaptation method (${\rm CA}^{2}$). We generalize the center-based method to unknown classes and optimize this objective based on the prior existing in the pre-trained network, i.e., pre-trained features that belong to the same class are adjacent. ${\rm CA}^{2}$ is validated to consistently improve OCC performance across a spectrum of training data classes, spanning from 1 to 1024, outperforming current state-of-the-art methods. Code is available at \href{https://github.com/zhangzilongc/CA2}{https://github.com/zhangzilongc/CA2}.
\end{abstract}

\section{Introduction}

One-class classification (OCC) refers to defining a boundary around the target classes, such that it accepts as many of the target objects as possible while minimizing the chance of accepting outlier objects. The effectiveness of classic OCC methods (like SVDD \cite{tax2004support}), applied to the low dimensional features, has been verified. However, when the dimension of features increases, these classic methods generally fail due to insufficient computational scalability and the curse of dimensionality. The emergence of deep learning provides an effective way to transform high-dimensional data into low-dimensional features, which makes it possible to re-apply the classic OCC methods. A subsequent problem is: how to use deep learning to learn a transformation in OCC? Compared to traditional supervised representation learning in multi-class classification, OCC only has information for a "single" class, making conventional supervised learning methods ineffective.

To learn representations in OCC, existing methods used self-supervised learning, where the supervised information of the learning objective derives from one-class samples themselves. However, \cite{deecke2021transfer} discovered that learning representations from scratch using some self-supervised learning strategies resulted in low-level feature learning, making it challenging to generalize to unseen semantics. To improve the discrimination to the unseen semantics, many previous works relied on pre-trained networks containing rich semantic information. \cite{bergman2020deep} revealed that utilizing the pre-trained features to detect outliers outperforms many OCC methods trained from scratch. To further improve the discrimination of pre-trained features on the target dataset, current methods adapted pre-trained features to eliminate the domain gap between the target dataset and the pre-trained dataset. 

Existing fine-tuning methods can be mainly divided into four classes: 1.) Center-based. These methods advocated adapting the features towards the center of one-class samples. The prior assumption is that all one-class samples belong to a single class. 2.) Contrastive-learning based. These methods stressed the alignment among the positive samples and the repulsion between the positive and negative samples. The prior assumption is that the one-class samples belong to a large number of classes. 3.) Outlier-exposure based. These methods generated or borrowed an external dataset as outliers, then accomplish a binary classification between one-class samples and outliers. The prior assumption is that the real unseen outliers are similar to the external dataset. Such an assumption becomes unrealistic as the categories of outliers increase. 4.) Pseudo-label based. These methods utilized pseudo labels created by clustering algorithms to fine-tune features. The prior assumption is that the number of categories needs to be known in advance, and the number of categories should not be too large. 

In real scenarios, the classes of one-class samples and outliers are generally unknown. For example, online tools such as iNaturalist (2019) enable photo-based classification and subsequent cataloging in data repositories from pictures uploaded by naturalists \cite{ahmed2020detecting}. A request for such a tool is to identify the novel species. Since the data are heterogeneous in repositories, one-class samples belong to multiple classes; In video anomaly detection, normal (one-class) samples may belong to the same class or diverse classes due to different perspectives, monitoring areas, and human behavior. In other words, the assumptions of current fine-tuning methods more or less do not match the real scenarios, which degrade the discrimination of original pre-trained features.

In this paper, we propose a simple \textbf{c}lass-\textbf{a}gnostic \textbf{a}daptive feature adaptation method (${\rm CA}^{2}$). We start from a center-based method and then generalize this objective to unknown classes. ${\rm CA}^{2}$ hinges on a nature advantage of the pre-trained model, i.e., pre-trained features that belong to the same class are adjacent. With this prior, ${\rm CA}^{2}$ adaptively clusters the features of every class tighter. To validate the effectiveness of ${\rm CA}^{2}$, we perform the experiments, varying the number of one-class sample classes from 1 to 1024. The results show that ${\rm CA}^{2}$ can achieve consistent improvement, while the current state-of-the-art (SOTA) methods fail in some cases. \textbf{Our contributions are following:}

\begin{enumerate}
	\item We perform experiments across a spectrum of training classes from 1 to 1024 and empirically show the limitations of current SOTA methods.
	\item We propose a simple class-agnostic feature adaptation method. It can achieve a consistent improvement under different numbers of classes and surpass SOTA methods in some cases.
\end{enumerate}

\section{Related Work}

Fine-tuning the pre-trained features on the downstream task has been a paradigm for improving the performance of out-of-distribution (OOD) detection or OCC \cite{fort2021exploring, kumar2022fine}. Compared with the supervised fine-tuning in OOD, the feature adaptation in OCC is unsupervised, and the information (e.g., class num) of the data is unknown. The works for adapting the pre-trained features in the context of OCC can be approximately classified into four classes.

\noindent{\textbf{Center-based}} The inspiration for these works is from Support Vector Data Description (SVDD) \cite{tax2004support}, where the objective is to find the smallest hypersphere that encloses the majority of the features. \cite{ruff2018deep} generalized this idea to learn a neural network to map most of the representations into a hypersphere. \cite{reiss2021panda, perera2019learning} further utilized this idea to fine-tune the pre-trained features. The central thought of these works is to fine-tune the features towards a certain point so that the features of one-class samples can be more compact. Essentially, the assumption underlying this idea is that all one-class samples are from the same class.

\noindent{\textbf{Contrastive-learning based}} Contrastive learning (CL) \cite{hadsell2006dimensionality} is one of the mainstream methods for unsupervised representation learning, where the core idea is to minimize the distance between positive pairs while maximizing that of negative ones. Since the unsupervised scenario in CL is consistent with that in OCC, it naturally becomes a method to learn the features \cite{tack2020csi, sohn2020learning}. However, \cite{reiss2023mean} showed that fine-tuning pre-trained features with CL degraded the original discrimination, and then they proposed mean-shift CL to alleviate the degradation. Essentially, the degradation of CL-based methods comes from maximizing the distance of negative pairs. In situations where the one-class samples are concentrated within a few classes, the repulsion between pairs of negative samples is more likely to involve samples from the same class. This phenomenon contradicts the principle of compactness that is intended. In other words, the assumption of adapting the features with CL-based is that one-class samples are from a large number of categories.

\noindent{\textbf{Outlier-exposure based}} The core idea of these methods is to formulate unsupervised feature adaptation as a supervised binary classification. These methods \cite{deecke2021transfer, liznerski2020explainable} borrowed an external dataset as the outliers and then fine-tuned features to distinguish the outliers. However, \cite{ye2021understanding} pointed out that the relevance between the external dataset and real unseen outliers greatly affects OCC performance. Recent method \cite{mirzaei2023fake} proposed to generate the outliers based on training data. However, it is challenging to apply this method to training data containing unknown categories since the hyperparameters of the outlier generation are sensitive to training data variations.

\noindent{\textbf{Pseudo-label based}} These methods \cite{cohen2022out} followed the idea of fine-tuning in supervised learning. The practice is to generate the pseudo labels by some unsupervised clustering methods \cite{van2020scan} and then use a cross-entropy loss to fine-tune the pseudo labels. The weakness is that these methods relied on the preset number of labels.

In real scenarios, the information of the one-class samples is unknown, leading to the inapplicability of existing adaptation methods. To tackle this issue, we propose a simple class-agnostic adaptive feature adaptation method.

\section{Method}

\subsection{Problem Setup}

In OCC, we are given a set of one-class samples without outliers, aiming to classify a new sample $\textbf{x}$ as being a one-class sample or an outlier. The method used in this paper for tacking OCC follows the standard two-stage setting in \cite{li2021cutpaste, reiss2023mean}, where the first step is to learn a low-dimensional representation of a sample by the neural network function $f_{\theta}(\cdot) \in \textbf{R}^{d}$. Subsequently, in the second stage, the sample is determined as being a one-class sample or an outlier by the OCC classifier $s(\cdot ) \in R$. The score of a sample classified as an outlier is $s(f_{\theta}(\textbf{x}))$.

In the context of the feature adaptation of OCC, $f_{\theta}(\cdot)$ is initialized by pre-trained weights $f_{0}(\cdot)$, which can be learned using external datasets (e.g., ImageNet \cite{russakovsky2015imagenet}). The main goal in such a setting is to further improve the discrimination between one-class samples and unseen outliers by fine-tuning $f_{0}(\cdot)$. Note that during the feature adaptation, learning is unsupervised. Only in the second step, all training samples are assigned the "one" category.

\begin{figure}[!t]
\centering
\includegraphics[width=3.2in]{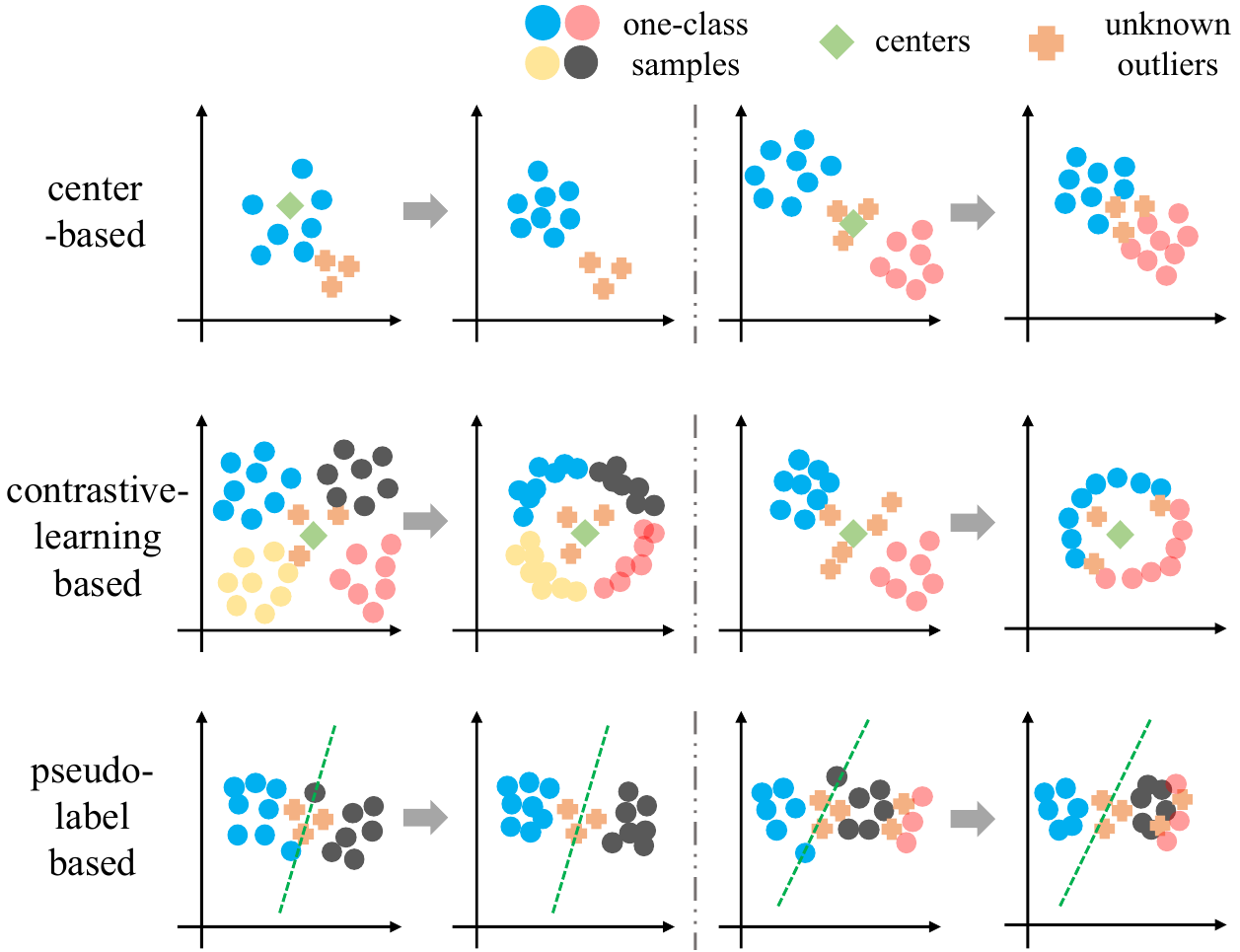}
\caption{\textbf{Top:} Center-based methods cluster the samples towards the center of samples. Left: It works well as the data belong to the same class. Right: But when the samples belong to multiple classes, this degrades the initial discrimination. \textbf{Middle:} Contrastive-learning based methods can cluster the samples around the center in a uniform distribution. Left: When the number of categories belongs to a large number of categories, this objective can achieve an ideal result. Right: But when there are few categories, this objective may confuse the one-class features with the outliers. \textbf{Bottom:} Pseudo-label based methods optimize a standard cross entropy loss based on the generated pseudo labels. Left: When the number of preset clusters is close to the number of real categories, they can obtain a good result. Right: But when the number of preset clusters is inconsistent, this objective degrades the original discrimination.}
\label{fig_1}
\end{figure}

\subsection{Preliminary}

Before elaborating on our method, we first review the current SOTA methods and analyze their limitations when applied to scenarios involving varying numbers of classes.

\subsubsection{Center-based} The center-based methods inspired by the idea of classic SVDD proposed the following optimization objective:

\begin{equation}
\label{EN:1}
\min_{\boldsymbol{\theta}} \mathop{{\rm \mathbb{E}}}_{\textbf{x}_{i}} ||f_{\theta}(\textbf{x}_{i})-\textbf{c}||_{2}^2+\upsilon R(\boldsymbol{\theta}),
\end{equation}

\noindent where \textbf{c} is the mean of representations of one-class samples (a constant vector), and $R(\cdot)$ denotes a regularization term to avoid the collapse of features. The idea is to fine-tune the features towards the fixed center so that the features of one-class samples can be more compact. Obviously, such an objective will degrade the discrimination when the one-class samples belong to multiple classes, as shown in Fig. \ref{fig_1} Top.

\subsubsection{Contrastive-learning based} \cite{reiss2023mean} proposed mean-shift contrastive learning (CL) to alleviate the degradation caused by CL. The objective is as follows:

\begin{equation}
\label{EN: add1}
\begin{split}
& \min_{f} \mathop{{\rm \mathbb{E}}}_{\textbf{x}_{i}} -{\rm log} ( \\
& \frac{{\rm exp}({\rm sim}(\frac{f(\hat{\textbf{x}}_{i_{1}})}{||f(\hat{\textbf{x}}_{i_{1}})||}-\textbf{c}^{*}, \frac{f(\hat{\textbf{x}}_{i_{2}})} {||f(\hat{\textbf{x}}_{i_{2}})||}-\textbf{c}^{*})/\tau)}{\Sigma_{m=1}^{2B}1_{[m\neq i]}{\rm exp}({\rm sim}(\frac{f(\hat{\textbf{x}}_{i_{1}})}{||f(\hat{\textbf{x}}_{i_{1}})||}-\textbf{c}^{*}, \frac{f(\hat{\textbf{x}}_{m_{1}})} {||f(\hat{\textbf{x}}_{m_{1}})||}-\textbf{c}^{*})/\tau)}),
\end{split}
\end{equation}

\noindent where $\hat{\textbf{x}}_{i_{1}}$ and $\hat{\textbf{x}}_{i_{2}}$ are two different data augmentations of $\textbf{{x}}_{i}$, $\hat{\textbf{x}}_{m_{1}}$ refers to samples in the same batch as $\hat{\textbf{x}}_{i_{1}}$, $\tau$ denotes a temperature, $\textbf{c}^{*}$ denotes the center of the normalized feature representations and ${\rm sim}(\cdot)$ denotes the cosine similarity. This objective forces features to be distributed around the center in a uniform distribution. As shown in Fig. \ref{fig_1} middle, when the number of sample categories belongs to a large number of categories, this objective can achieve an ideal result. However, when there are few categories, this objective may confuse the one-class features with the outliers.

\subsubsection{Pseudo-label based} These methods \cite{cohen2022out} relied on the pseudo labels generated by some unsupervised clustering methods and then optimized a standard cross-entropy loss based on the pseudo labels. As shown in Fig. \ref{fig_1} bottom, when the number of preset clusters is close to the number of real categories, they can obtain a good result. But when the number of preset clusters is much larger or smaller than the real value, this objective degrades the original discrimination. We will empirically illustrate this in the experiments.

\begin{figure}[!t]
\centering
\includegraphics[width=3.3in]{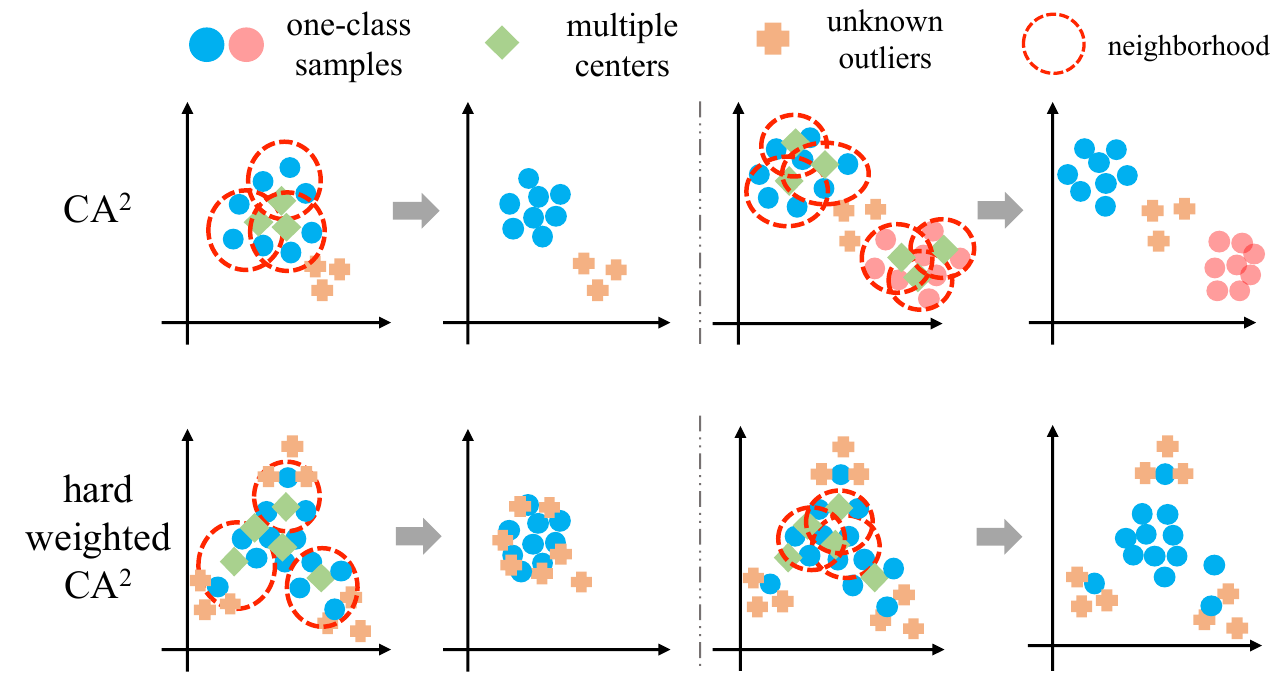}
\caption{\textbf{Top:} Mechanism of ${\rm CA}^{2}$: ${\rm CA}^{2}$ forces every sample to cluster towards the mean (center) of their $k$ nearest neighbors. Since the pre-trained features belonging to the same category are adjacent, the features of each category will be clustered more closely. \textbf{Bottom:} Left (failure mode): As there is a large overlapping between one-class samples and unseen outliers, clustering the samples at the edge of a certain class towards the center degrades the original discrimination. Right: Hard weighted ${\rm CA}^{2}$ only clusters the samples near the class center to alleviate this issue.}
\label{fig_2}
\end{figure}

\subsection{$\boldsymbol{{\rm CA}^{2}}$}

To propose a class-agnostic adaptive feature adaptation method, we start to generalize Eq. (\ref{EN:1}) to multiple classes. Suppose the training data belongs to $m$ classes ($m$ can be 1), and the center of every class is $\textbf{c}_{m}$. The optimization objective is following:

\begin{equation}
\label{EN:2}
\min_{\boldsymbol{\theta}} \mathop{{\rm \mathbb{E}}}_{m} \mathop{{\rm \mathbb{E}}}_{\textbf{x}_{i}\in C_{m}} ||f_{\theta}(\textbf{x}_{i})-(\textbf{c}_{m}+\boldsymbol{\varepsilon})||_{2}^2 +\upsilon R(\boldsymbol{\theta}),
\end{equation}

\noindent where $C_{m}$ is the set that $\textbf{x}_{i}$ belongs to the m-th class and $\boldsymbol{\varepsilon}$ is the noise where $||\boldsymbol{\varepsilon}||_{2}$ is under a controllable threshold. Obviously, when $m=1$, Eq. (\ref{EN:1})  is a special case of Eq. (\ref{EN:2}). Note that for the new training data, we do not know how many categories (i.e., $m$) are in the training set and which class each sample belongs to. These conditions make it intractable to solve Eq. (\ref{EN:2}). 

In this paper, we propose an ingenious method to bypass the above problem. Specifically, since the initial features are from a pre-trained network, \textit{the encoded features of different categories are discriminative, while the features belonging to the same category are adjacent.} Relying on this prior, we can replace $\textbf{c}_{m}+\boldsymbol{\varepsilon}$ in Eq. (\ref{EN:2}) with a mean of $k$ nearest neighbors of $\textbf{x}_{i}$. The optimization objective is following:

\begin{equation}
\label{EN:3}
\min_{\boldsymbol{\theta}} \mathop{{\rm \mathbb{E}}}_{\textbf{x}_{i}} ||f_{\theta}(\textbf{x}_{i})-\kappa(\textbf{x}_{i})||_{2}^2 +\upsilon R(\boldsymbol{\theta}),
\end{equation}

\noindent where $\kappa(\cdot)$ denotes a operater that represents a mean of $k$ nearest neighbors of $\textbf{x}_{i}$. The process of Eq. (\ref{EN:3}) is described in Fig. \ref{fig_2} Top. For samples within the same class, since the features are distributed around the centers of features from more to less, this ultimately leads to a more dynamic and compact clustering of features for each class.


For the regularization term $R(\cdot)$, the previous works \cite{reiss2021panda, ruff2018deep} used early stopping, gradient clipping, or bias removing to alleviate the collapse of features. In our study, we refrain from employing these techniques. Instead, the inherent clustering of each sample towards the centers of their $k$ nearest neighbors serves as a built-in regularization, safeguarding against the risk of the representation converging to a fixed point.

\begin{algorithm}[t]
\small{
\caption{${\rm CA}^{2}$}
\label{alg: algorithm}
\begin{algorithmic}[1]
\State \textbf{Input:} Dataset $\mathcal{D}$, Pre-trained Neural Net $f_\theta$, Neighbors $\mathcal{N}_\mathcal{D}=\{\}$, Distance $\mathcal{P}=\{\}$, Hyper-parameter $\beta, k$.
\For{$x_i \in \mathcal{D}$}
\State $\mathcal{N}_\mathcal{D} \leftarrow \mathcal{N}_{\mathcal{D}} \cup \mathcal{N}_{x_i}$, with $\mathcal{N}_{x_i} = k$ neighboring samples of $f_\theta(x_i)$. 
\State $\mathcal{P} \leftarrow \mathcal{P} \cup \tau_{i}$, with $\tau_{i}={f_{\theta}^{*}(\textbf{x}_{i})}^{\rm{T}}\kappa^{*}(\textbf{x}_{i})$. 
\EndFor

\State Obtain a mean distance $\tau$ and dynamic threshold $\sigma$.

\While{${\rm CA}^{2}$-loss decreases}
\State obtain a weight $w_{i}$ by Eq.~(\ref{EN: 4})
\State Update $f_\theta$ with Eq.~(\ref{EN:5})
\EndWhile 

\State \textbf{Return:} $f_\theta$
\end{algorithmic}
}
\end{algorithm}

\begin{table*}
\scriptsize
    \centering
    \begin{tabular}{c c c c c c c c c c c c}
    \hline
    \diagbox{method}{class num} & $2^1$ & $2^2$ & $2^3$ & $2^4$ & $2^5$ & $2^6$ & $2^7$ & $2^8$ & $2^9$ & $2^{10}$ & mean \\ \hline
    & $\uparrow$ / $ \downarrow$ & $\uparrow$ / $ \downarrow$ & $\uparrow$ / $ \downarrow$ & $\uparrow$ / $ \downarrow$ & $\uparrow$ / $ \downarrow$ & $\uparrow$ / $ \downarrow$ & $\uparrow$ / $ \downarrow$ & $\uparrow$ / $ \downarrow$ & $\uparrow$ / $ \downarrow$ & $\uparrow$ / $ \downarrow$ & $\uparrow$ / $ \downarrow$ \\ \hline
    baseline & 97.4/4.4 & 94.6/30.0 & 96.0/24.8 & 94.2/37.7 & 91.2/51.6 & 88.5/58.2 & 85.6/64.2 & 83.3/68.6 & 78.4/76.5 & 75.6/81.0 & 88.5/49.7 \\ 
    PANDA {\fontsize{5pt}{2pt}\selectfont CVPR' 2021} & 97.5/4.1 & 94.7/{\color{red}30.2} & 96.2/23.4 & 94.3/37.7 & 91.4/46.6 & 88.7/54.6 & 85.7/63.0 & {\color{red}82.9}/65.5 & {\color{red}78.3}/73.4 & {\color{red}75.3}/76.4 & 88.5/47.1 \\
    ADIB {\fontsize{5pt}{2pt}\selectfont ICML' 2021} & 97.6/\textbf{0.5} & 95.7/\textbf{25.2} & \textbf{98.0}/\textbf{7.7} & \textbf{95.8}/\textbf{25.3} & 91.7/47.9 & 89.2/{\color{red}58.5} & 85.6/{\color{red}65.6} & {\color{red}83.0}/{\color{red}70.1} & {\color{red}78.1}/{\color{red}79.2} & {\color{red}75.2}/{\color{red}83.0} & 89.0/46.3 \\
    OODWCL {\fontsize{5pt}{2pt}\selectfont ECCV W' 2022} & {\color{red}97.2}/4.1 & 94.9/{\color{red}33.3} & 96.3/21.8 & 95.3/30.9 & 92.2/43.6 & 89.5/50.0 & 87.4/55.8 & 85.0/61.8 & 79.8/71.7 & {\color{red}73.5}/{\color{red}84.6} & 89.1/45.8 \\
    MSCL {\fontsize{5pt}{2pt}\selectfont AAAI' 2023} & 97.4/4.3 & 94.6/28.9 & 96.3/22.6 & 94.6/38.7 & 91.8/45.8 & 89.4/52.4 & 86.5/59.6 & 83.7/62.8 & 79.7/71.3 & 77.5/73.9 & 89.2/46.0 \\
    ${\rm CA}^{2}$ (\textbf{Ours}) & \textbf{97.9}/0.9 & \textbf{95.8}/26.1 & 97.0/16.9 & 95.3/33.8 & \textbf{92.5}/\textbf{42.4} & \textbf{90.6}/\textbf{46.6} & \textbf{87.9}/\textbf{55.6} & \textbf{85.4}/\textbf{56.2} & \textbf{80.7}/\textbf{66.2} & \textbf{78.6}/\textbf{68.9} & \textbf{90.2}/\textbf{41.4} \\\hline
    \end{tabular}
     \caption{One-class classification performance (AUROC \% $\uparrow$ / TPR95FPR \% $\downarrow$) under different numbers of classes. Bold denotes the best results. {\color{red}Red} indicates that the result has decreased compared to the initial value. The results over 3 trials are reported.}
    \label{Ta1}
\end{table*}

\subsection{Failure Mode}
\label{sec: 1}

As shown in Fig. 2 Bottom, when there is a large overlapping between one-class samples and unseen outliers, clustering the samples at the edge of a certain class towards the center may degrade the original discrimination. This is very common in near-distribution detection \cite{mirzaei2023fake}. To alleviate this issue, we propose a hard weighted ${\rm CA}^{2}$. Specifically, we endow the samples near the center with a bigger weight $w_{i}$ than the samples at the edge. We simply use the distance $||f_{\theta}(\textbf{x}_{i})-\kappa(\textbf{x}_{i})||_{2}^2$ to identify the samples near the center or at the edge. 

In our implementation, we endow 1 for the samples near the center and 0 for the samples at the edge. We use a $\rm{L}_{2}$ normalized feature of $f_{\theta}(\textbf{x}_{i})$ and $\kappa(\textbf{x}_{i})$ in Eq. (\ref{EN:3}), i.e., $f^{*}(\textbf{x}_{i})$ and $\kappa^{*}(\textbf{x}_{i})$. In the initial training stage, we obtain a mean distance $\tau$ of all training samples, i.e., $\tau=\mathop{{\rm \mathbb{E}}}_{\textbf{x}_{i}}{f_{\theta}^{*}(\textbf{x}_{i})}^{\rm{T}}\kappa^{*}(\textbf{x}_{i})$. Subsequently, a dynamic threshold, $\sigma=\tau+\beta(1-\tau)$, is derived to identify the samples near the center or at the edge, where $\beta\in [0, 1]$ is a hyper-parameter. The weight $w_{i}$ of every sample is:

\begin{equation}
w_{i}=\left\{
\begin{aligned}
1 & , & {f_{\theta}^{*}(\textbf{x}_{i})}^{\rm{T}}\kappa^{*}(\textbf{x}_{i}) > \sigma, \\
0 & , & {f_{\theta}^{*}(\textbf{x}_{i})}^{\rm{T}}\kappa^{*}(\textbf{x}_{i}) \leq \sigma.
\end{aligned}
\right.
\label{EN: 4}
\end{equation}

\noindent The final optimized objective is:

\begin{equation}
\label{EN:5}
\min_{\boldsymbol{\theta}} \mathop{{\rm \mathbb{E}}}_{\textbf{x}_{i}} -w_{i}{f^{*}(\textbf{x}_{i})}^{\rm{T}}\kappa^{*}(\textbf{x}_{i}).
\end{equation}

\noindent Algorithm \ref{alg: algorithm} summarizes all the steps of hard weighted ${\rm CA}^{2}$. In the subsequent parts, ${\rm CA}^{2}$ denotes hard weighted ${\rm CA}^{2}$.

\subsection{OCC Classifier}

In this paper, we utilize $k$-nearest neighbors (k-NN) \cite{eskin2002geometric} to obtain the score of a sample belonging to the outlier. We extract the feature of the test sample $f(\textbf{x})$ and search its $k$ nearest neighbors in the training set. Then we calculate the following index to obtain the score:

 \begin{equation}
\label{EN:6}
	\frac{1}{k} \mathop{{\rm \mathbb{\sum}}}_{\textbf{t}\in N_{k^*}(f(\textbf{x}))} || \textbf{t} - f(\textbf{x}) ||^2
\end{equation}

\noindent where $N_{k^*}(f(\textbf{x}))$ denotes the set of $k^*$ nearest neighbors of $f(\textbf{x})$.

\section{Experiments}

\subsection{Experimental Setups}

\subsubsection{Implementation Details}

In our experiments, we use a ResNet 50 \cite{he2016deep} pre-trained on ImageNet-1K \cite{russakovsky2015imagenet} (unless speciﬁed otherwise). The learning rate is $3e^{-4}$. The optimizer is SGD, the weight decay is $1e^{-3}$, the momentum is 0.9, and the batch size is 512. $k=5$ in $\kappa(\cdot)$, $k^{*}=2$ and $\beta=0.3$. We train the neural network until the loss no longer drops. The input image is resized to $256 \times 256$ and then center croped to $224 \times 224$.

\subsubsection{Evaluation Metric}

We adopt the area under the receiver operating characteristic curve (AUROC), which is widely adopted. In addition, we employ the false positive rate of outliers when the true positive rate of one-class examples is at 95\% (TPR95FPR).

\subsubsection{Methods of Comparison}

1) Baseline \cite{bergman2020deep} used a k-NN OCC classifier to evaluate performance on the pre-trained features. 2) PANDA \cite{reiss2021panda} used a center-based method to fine-tune the features. 3) OODWCL \cite{cohen2022out} used a pseudo-label based method to fine-tune features. 4) MSCL \cite{reiss2023mean} used mean-shift contrastive learning to fine-tune features. 5) ADIB \cite{deecke2021transfer} used an external dataset as outliers to fine-tune features. 6) FITYMI \cite{mirzaei2023fake} used a generated outlier dataset to fine-tune features. Since some models have the phenomenon of catastrophic forgetting, we report the best performance achieved by these methods during training. The evaluated initial backbone is a ResNet 50 pre-trained on ImageNet-1K (unless speciﬁed otherwise). More details are listed in the Supplementary Material (SM).

\subsection{Multi-class Experiments}

\subsubsection{Dataset} 

In the previous references, the maximum number of categories contained in the training data in OCC is 30 categories \cite{cohen2022out}. In this subsection, to validate the effectiveness of ${\rm CA}^{2}$, we perform the experiments with the number of one-class sample classes ranging from 2 to 1024, where the maximum number of categories is more than 30 times the current maximum. A large semantic space of training data, resulting in a large space of uncertainty, would increase the difficulty of outlier detection \cite{huang2021mos}. We use the part data proposed by \cite{zhang2022benchmarking} as one-class samples, where there is a similar semantic space between the used samples and ImageNet-1K. We extract the samples belonging to diverse classes as one-class samples respectively (ranging from 2 to 1024). In addition, in the previous references, the evaluated unseen outliers more or less \textit{overlap with the semantic space of the pre-trained model}, i.e., the semantic space in ImageNet-1K includes the semantic space of unseen outliers. This is contrary to the goal of OCC. For tackling this issue, we use an outlier dataset of ImageNet-1K proposed by \cite{bitterwolf2023or}, where there are 64 OOD classes and no ImageNet-1K class objects.

\begin{figure}[!t]
\centering
\includegraphics[width=3.2in]{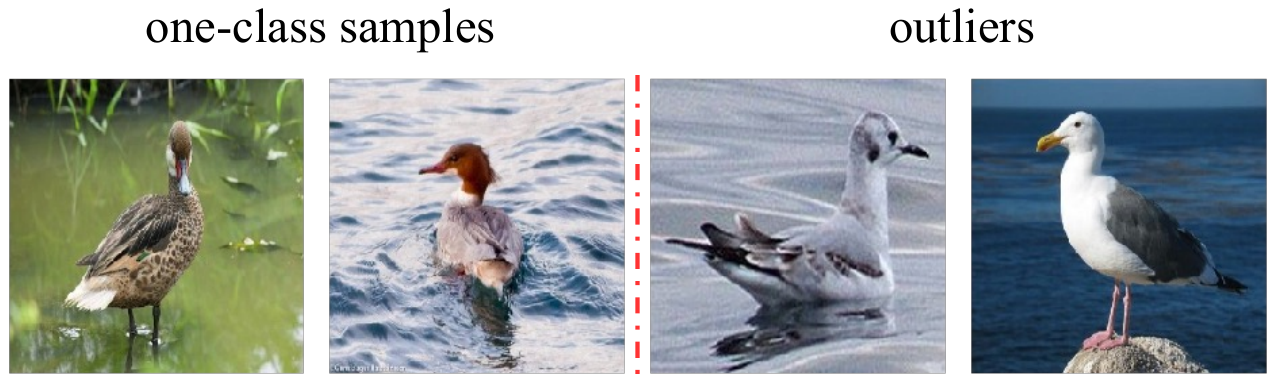}
\caption{Some samples in the near distribution experiment.}
\label{fig_3}
\end{figure}

\subsubsection{Results}

The results are listed in Table \ref{Ta1}. 1) When the outliers remain unaltered, the baseline performance generally decreases as the number of categories increases. This observation substantiates the previous conclusion that a large semantic space results in a large space of uncertainty. 2) PANDA: When the one-class samples belong to multiple classes (especially class num $>2^7$), AUROC after adaptation is lower than the initial value. Note that since we report the best performance during training, the best AUROC (class num $\leq2^7$) is slightly higher than the initial value. It actually initially improves slightly and then drops off rapidly. ADIB: Even though the borrowed external dataset exhibits only a marginal correlation with the real outliers, ADIB can still achieve a favorable improvement provided a powerful initial discrimination between the one-class samples and outliers. However, if the initial discrimination below a certain threshold ($>2^6$), ADIB begins to degrade at the beginning of training. This stems from the fact that the external dataset is not similar to the real outliers, leading to confusion between features in fine-tuning. In other words, the similarity between the external dataset and real outliers only works when the initial discrimination is below a certain threshold. This is a supplement to the existing conclusion \cite{ye2021understanding}. OODWCL: Given an inconsistent preset clustering number, the feature adaptation results in a degradation of the original discrimination whether the underlying classes are $2^1$ or $2^{10}$. We have determined that this primarily stems from errors introduced during the pseudo-label generation stage. This issue becomes more pronounced especially when dealing with preset numbers of clusters that differ significantly from real categories. MSCL: Although there is a consistent improvement for the best performance, when the number of categories is few, the improvement is comparatively subdued and the decline occurs more rapidly. This alignment with expectations can be attributed to the repulsion among intra-class samples, which is particularly prominent when dealing with a smaller number of classes. 3) ${\rm CA}^{2}$ yields a considerable boost for diverse categories. Especially when the number of categories is large ($>2^5$), it surpasses the current SOTA method.

\subsection{Single Class Experiments}

\subsubsection{Dataset}

We evaluate the effectiveness of ${\rm CA}^{2}$ on a near distribution experiment and a common CIFAR 10 one vs rest, where the one-class samples belong to the single preset semantics. Near distribution experiment: we extract the samples that have the same coarse-grained semantics but different fine-grained semantics from the DyML-Animal dataset \cite{sun2021dynamic}. Some examples are shown in Fig. 3. CIFAR 10 one vs rest: Following the standard protocol \cite{sohn2020learning, tack2020csi, reiss2021panda}, the images from one class in CIFAR 10 \cite{krizhevsky2009learning} are given as one-class samples, and those from the remaining classes are given as outliers. For a dataset including $c$ classes, we perform $c$ experiments and report mean AUROC. In this dataset, we use the common pre-trained ResNet 152.

\subsubsection{Results}

The results are listed in Table \ref{Ta2} and \ref{Ta3}. Compared with Table \ref{Ta1}, there is a large boost for PANDA. The reason is that the class num of data is consistent with the assumption of PANDA. In the case of MSCL, although it can obtain a considerable improvement on the benchmark, its improvement is weak in the near distribution experiment. The main reason is that the semantics in the near distribution experiment are finer than the semantics of the single class in the benchmark. For example, the “bird” class in CIFAR 10 contains different breeds of bird. The “bird” class is coarse-grained. This results in a weaker repulsion among positive samples. However, when confronted with the fine-grained semantics of the near distribution experiment, the heightened repulsion among positive samples becomes pronounced, ultimately resulting in weaker performance. For ${\rm CA}^{2}$, no matter in the near distribution experiment or benchmark, ${\rm CA}^{2}$ can obtain a stable boost. Notably, the performance of ${\rm CA}^{2}$ is similar to that of PANDA. This further illustrates that PANDA is a special case of ${\rm CA}^{2}$.

\begin{table}
\scriptsize
    \centering
    \begin{tabular}{c c c c c c}
    \hline
    baseline & PANDA & ADIB & OODWCL & MSCL & \textbf{Ours} \\ \hline
    78.6 & 84.6 & {\color{red}78.3} & 78.8 & 79.8 & \textbf{84.8} \\ \hline
    \end{tabular}
    \caption{AUROC (\%) on the near distribution experiment.}
    \label{Ta2}
\end{table}

\begin{table}
\scriptsize
    \centering
    \begin{tabular}{c c c c c c c}
    \hline
   baseline & PANDA & FITYMI & \textbf{Ours} & baseline* & MSCL* & \textbf{Ours}* \\ \hline
   92.5 & 96.2 & 95.3 & 96.1 & 95.8 & \textbf{97.2} & \textbf{97.2} \\ \hline
    \end{tabular}
    \caption{mean AUROC (\%) on CIFAR 10 one vs rest. * denotes that the features are $\rm{L}_{2}$ normalized.}
    \label{Ta3}
\end{table}

\begin{figure}[!t]
\centering
\includegraphics[width=3.3in]{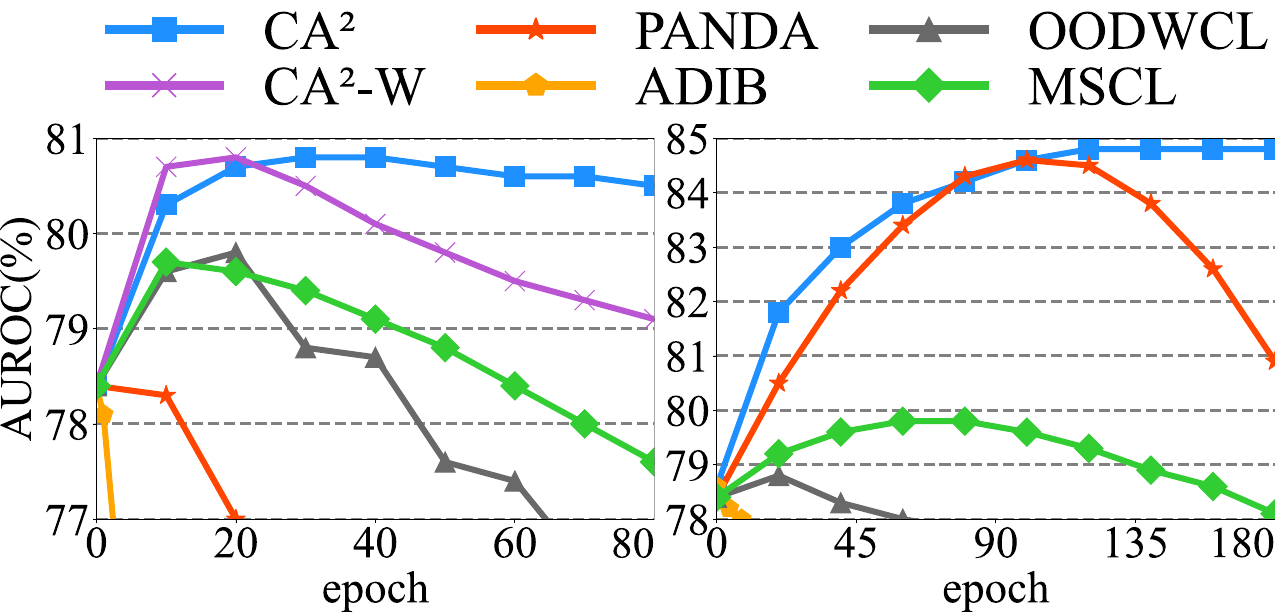}
\caption{AUROC variation of different methods. ${\rm CA}^{2}-{\rm W}$ denotes ${\rm CA}^{2}$ without dynamic weights. \textbf{Left}: multi-class experiment with 512 categories. \textbf{Right}: near distribution experiment with the single class.}
\label{fig_ab_1}
\end{figure}

\section{Ablations}

In this section, we ablate some factors closely related to ${\rm CA}^{2}$. The multi-class data with 512 categories is the default experimental dataset. The other setups follow the default. 

\subsection{Dynamic Weight}

In this subsection, we investigate the effect caused by dynamic weight $w_{i}$. We firstly remove the weight $w_{i}$, i.e., regard the weight of every sample as 1 in Eq. (\ref{EN:5}). We name this version as ${\rm CA}^{2}-{\rm W}$. The result is shown in Fig. \ref{fig_ab_1} Left. The result reveals that as the performance of ${\rm CA}^{2}-{\rm W}$ reaches its peak, its performance begins to decline rapidly compared with ${\rm CA}^{2}$. The main reason is that as the training progresses, the overlapping between one class samples and outliers is exacerbated. This is essentially the samples at the edge getting closer to the center, as described in the failure mode of Method. Using a hard weighted ${\rm CA}^{2}$ can alleviate this.

We also report the effect caused by different $\beta$, as listed in Table \ref{Ta4}. We find that when $\beta$ is small, the performance first increases and then decreases, which is similar to the above phenomenon. When $\beta$ is large, tightening towards the center only works on a few features, resulting in a large number of features unchanged. The improvement is also poor.

\begin{table}
\footnotesize
    \centering
    \begin{tabular}{c c c c c}
    \hline
   baseline & $\beta=0$ & $\beta=0.3$ & $\beta=0.6$ & $\beta=0.9$ \\ \hline
   78.4 & 79.1 & 80.7 & 80.2 & 78.9 \\ \hline
    \end{tabular}
    \caption{AUROC (\%) under different $\beta$.}
    \label{Ta4}
\end{table}

\begin{table}
\footnotesize
    \centering
    \begin{tabular}{c c c c c}
    \hline
   baseline & $k=2$ & $k=5$ & $k=10$ & $k=20$ \\ \hline
   78.4 & 80.6 & 80.7 & 80.4 & 79.7 \\ \hline
    \end{tabular}
    \caption{AUROC (\%) under different $k$.}
    \label{Ta5}
\end{table}

\subsection{Num of Nearest Neighbor}

The num of nearest neighbor $k$ in $\kappa(\cdot)$ affects the center of every sample. We report the results caused by different $k$ in Table \ref{Ta5}. When $k$ increases, the performance starts to decline. The primary rationale is that a large $k$ brings more noise to the edge samples. As a result, the substitution in Eq. (\ref{EN:2}) does not hold for some samples. 

\begin{table}
\scriptsize
    \centering
    \begin{tabular}{c c c c c}
    \hline
   & \multicolumn{2}{c}{\shortstack{Supervised Pre-training}} & \multicolumn{2}{c}{\shortstack{Self-supervised Pre-training}} \\
   & ResNet 152 & ViT Base & ResNet 50 & ViT Base \\ \hline
  initial & 79.0 & 86.2 & 66.4 & 80.4 \\ 
  adaptation & 81.7 & 88.6 & 66.7 & 82.7 \\ \hline
    \end{tabular}
    \caption{AUROC (\%) on different pre-trained models.}
    \label{Ta6}
\end{table}

\begin{figure}[!t]
\centering
\includegraphics[width=3.1in]{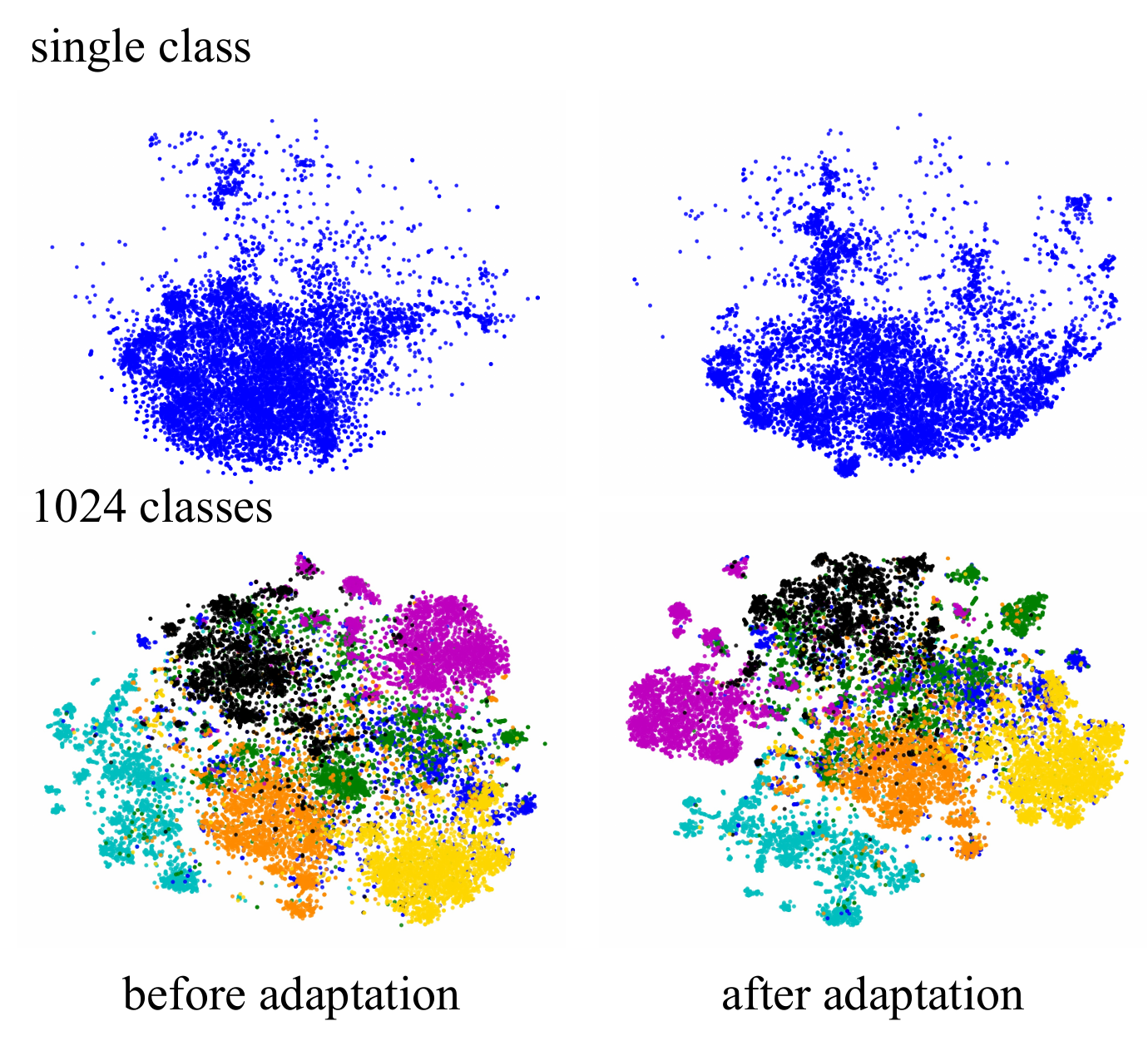}
\caption{T-SNE visualization before and after feature adaptation. \textbf{Top:} one-class samples in the near distribution experiment. \textbf{Bottom:} one-class samples in the multi-class experiment with 1024 categories. Points of the same color represent coarse-grained semantics (a total of 7).}
\label{fig_dis_1}
\end{figure}

\subsection{Pre-trained Models}

The mechanism of ${\rm CA}^{2}$ heavily hinges on the prior that the pre-trained features belonging to the same category are adjacent. In this subsection, we show the sensitivity of ${\rm CA}^{2}$ to different pre-trained Models. We show the results from the perspective of different layers, types of backbones, and pre-training. More details are described in SM. The results are shown in Table \ref{Ta6}. We find that ${\rm CA}^{2}$ is not sensitive to layers, types of backbones, and pre-training. However, its sensitivity is closely tied to the initial discrimination of pre-trained models. In cases where the initial discrimination is weak, the training process exhibits a pattern: initial improvement in discrimination followed by rapid deterioration. This mainly originates from the noise brought by the nearest neighbors. Weak initial discrimination leads to greater confusion of outliers and one-class samples.

\section{Discussion}

\subsection{Catastrophic Forgetting} 

Catastrophic forgetting (CF) denotes the phenomenon where performance initially improves but subsequently deteriorates as the training advances. CF has been a longstanding intractable problem in OCC \cite{ruff2018deep, reiss2023mean}. Given that real outliers are often inaccessible, rendering early training termination impractical for achieving optimal performance, the prevention or alleviation of CF becomes paramount. The essential origin of CF is that the outliers and one-class samples are gradually mixed. 

In this subsection, we show the AUROC variation of different methods in different experiments. The results are shown in Fig. \ref{fig_ab_1}. We can find that ${\rm CA}^{2}$ has a strong ability against the CF compared with the other methods. There are two main reasons: 1. Unlike PANDA, each sample has its own fixed center in ${\rm CA}^{2}$. This keeps the model from collapsing to a constant. In the near distribution experiment, ${\rm CA}^{2}$ does not degenerate like PANDA. 2. Compared with other methods, the features are adapted relative to the position of the original pre-trained features. This reduces the possibility of overlap with outliers due to feature shifting. This is similar to the recent supervised fine-tuning method \cite{kumar2022fine} in which the authors first freeze the pre-trained features to train the classifier, and then unfreeze the backbone for joint training. This has been shown to improve the capabilities of OOD. In addition, the hard weight $w_{i}$ is also a crucial element to alleviate CF.

\begin{table}
\footnotesize
    \centering
    \begin{tabular}{c c c c}
    \hline
  class num & 64 & 256 & 1024 \\ \hline
  initial & 87.1 & 71.6 & 48.9  \\ 
  adaptation & 87.9 & 72.0 & 49.2 \\ \hline
    \end{tabular}
    \caption{Linear classification (top-1 accuracy \%) before and after adaptation.}
    \label{Ta7}
\end{table}

\subsection{Mechanism of ${\rm CA}^{2}$}

The mechanism of ${\rm CA}^{2}$ is to force every sample to cluster towards the mean (center) of their $k$ nearest neighbors, which results in diverse tight "sub-class". We show t-SNE visualizations \cite{van2008visualizing} of features of the test set before and after adaptation. The results are shown in Fig. \ref{fig_dis_1}. As expected, there are more small groups (sub-class) after feature adaptation, both in multi-class and single-class experiments. In fact, in multi-class experiments, these small groups denote some fine-grained classes. This validates the mechanism of ${\rm CA}^{2}$.

\subsection{Unsupervised Feature Adaptation to Improve Linear Separability}

Improving the linear discriminative ability of the model for the in-distribution data and the ability to detect outliers are ideal targets for OOD detection \cite{du2022vos, kumar2022fine}. For OCC, achieving both intentions simultaneously is challenging due to the unavailable labels of in-distribution data. In addition, a large semantic space of training data exacerbates this difficulty. In this subsection, we show the ability of ${\rm CA}^{2}$ for improving linear separability under an unsupervised setup. We use a common linear probing \cite{he2020momentum, chen2020exploring} to evaluate the linear separability. We follow the steps in \cite{ericsson2021well}. The results are shown in Table \ref{Ta7}. The results show that ${\rm CA}^{2}$ can slightly improve the initial linear separability. Such a capability can benefit some unsupervised downstream tasks, e.g., unsupervised clustering \cite{van2020scan}, and unsupervised semantic segmentation \cite{melas2022deep}.

\section{Limitations}

1) ${\rm CA}^{2}$ is sensitive to the initial discrimination of pre-trained models. Weak initial discrimination makes $k$ nearest neighbors inaccurate. 2) A weakness of ${\rm CA}^{2}$ is a request for $k$ nearest neighbors searches before the training. When the size of the dataset is large (billion level), this will be extremely time-consuming.

\section{Conclusion}

This paper reveals the weakness of current feature adaptation methods when the number of categories of real-world training data is unknown. To tackle this issue, we generalize the classic idea of SVDD and propose a class-agnostic feature adaptation method. Although our method is simple, it can achieve a consistent improvement of OCC under different numbers of classes, which is more in line with the needs of real scenarios. Simple algorithms that scale well are the core of deep learning. While we focused on images, it is convenient to apply to other data modalities such as text.

\bibliography{ref.bib}

\begin{thebibliography}{34}
\providecommand{\natexlab}[1]{#1}

\bibitem[{Ahmed and Courville(2020)}]{ahmed2020detecting}
Ahmed, F.; and Courville, A. 2020.
\newblock Detecting semantic anomalies.
\newblock In \emph{Proceedings of the AAAI Conference on Artificial
  Intelligence}, volume~34, 3154--3162.

\bibitem[{Bergman, Cohen, and Hoshen(2020)}]{bergman2020deep}
Bergman, L.; Cohen, N.; and Hoshen, Y. 2020.
\newblock Deep nearest neighbor anomaly detection.
\newblock \emph{arXiv preprint arXiv:2002.10445}.

\bibitem[{Bitterwolf, M{\"u}ller, and Hein(2023)}]{bitterwolf2023or}
Bitterwolf, J.; M{\"u}ller, M.; and Hein, M. 2023.
\newblock In or Out? Fixing ImageNet Out-of-Distribution Detection Evaluation.
\newblock \emph{arXiv preprint arXiv:2306.00826}.

\bibitem[{Chen and He(2020)}]{chen2020exploring}
Chen, X.; and He, K. 2020.
\newblock Exploring Simple Siamese Representation Learning.
\newblock \emph{arXiv preprint arXiv:2011.10566}.

\bibitem[{Chen, Xie, and He(2021)}]{chen2021empirical}
Chen, X.; Xie, S.; and He, K. 2021.
\newblock An empirical study of training self-supervised vision transformers.
\newblock In \emph{Proceedings of the IEEE/CVF International Conference on
  Computer Vision}, 9640--9649.

\bibitem[{Cohen, Abutbul, and Hoshen(2022)}]{cohen2022out}
Cohen, N.; Abutbul, R.; and Hoshen, Y. 2022.
\newblock Out-of-Distribution Detection without Class Labels.
\newblock In \emph{European Conference on Computer Vision}, 101--117. Springer.

\bibitem[{Deecke et~al.(2021)Deecke, Ruff, Vandermeulen, and
  Bilen}]{deecke2021transfer}
Deecke, L.; Ruff, L.; Vandermeulen, R.~A.; and Bilen, H. 2021.
\newblock Transfer-based semantic anomaly detection.
\newblock In \emph{International Conference on Machine Learning}, 2546--2558.
  PMLR.

\bibitem[{Dosovitskiy et~al.(2020)Dosovitskiy, Beyer, Kolesnikov, Weissenborn,
  Zhai, Unterthiner, Dehghani, Minderer, Heigold, Gelly
  et~al.}]{dosovitskiy2020image}
Dosovitskiy, A.; Beyer, L.; Kolesnikov, A.; Weissenborn, D.; Zhai, X.;
  Unterthiner, T.; Dehghani, M.; Minderer, M.; Heigold, G.; Gelly, S.; et~al.
  2020.
\newblock An Image is Worth 16x16 Words: Transformers for Image Recognition at
  Scale.
\newblock In \emph{International Conference on Learning Representations}.

\bibitem[{Du et~al.(2022)Du, Wang, Cai, and Li}]{du2022vos}
Du, X.; Wang, Z.; Cai, M.; and Li, Y. 2022.
\newblock Vos: Learning what you don't know by virtual outlier synthesis.
\newblock \emph{arXiv preprint arXiv:2202.01197}.

\bibitem[{Ericsson, Gouk, and Hospedales(2021)}]{ericsson2021well}
Ericsson, L.; Gouk, H.; and Hospedales, T.~M. 2021.
\newblock How well do self-supervised models transfer?
\newblock In \emph{Proceedings of the IEEE/CVF Conference on Computer Vision
  and Pattern Recognition}, 5414--5423.

\bibitem[{Fort, Ren, and Lakshminarayanan(2021)}]{fort2021exploring}
Fort, S.; Ren, J.; and Lakshminarayanan, B. 2021.
\newblock Exploring the limits of out-of-distribution detection.
\newblock \emph{Advances in Neural Information Processing Systems}, 34:
  7068--7081.

\bibitem[{Hadsell, Chopra, and LeCun(2006)}]{hadsell2006dimensionality}
Hadsell, R.; Chopra, S.; and LeCun, Y. 2006.
\newblock Dimensionality reduction by learning an invariant mapping.
\newblock In \emph{2006 IEEE Computer Society Conference on Computer Vision and
  Pattern Recognition (CVPR'06)}, volume~2, 1735--1742. IEEE.

\bibitem[{He et~al.(2020)He, Fan, Wu, Xie, and Girshick}]{he2020momentum}
He, K.; Fan, H.; Wu, Y.; Xie, S.; and Girshick, R. 2020.
\newblock Momentum contrast for unsupervised visual representation learning.
\newblock In \emph{Proceedings of the IEEE/CVF Conference on Computer Vision
  and Pattern Recognition}, 9729--9738.

\bibitem[{He et~al.(2016)He, Zhang, Ren, and Sun}]{he2016deep}
He, K.; Zhang, X.; Ren, S.; and Sun, J. 2016.
\newblock Deep residual learning for image recognition.
\newblock In \emph{Proceedings of the IEEE conference on computer vision and
  pattern recognition}, 770--778.

\bibitem[{Huang and Li(2021)}]{huang2021mos}
Huang, R.; and Li, Y. 2021.
\newblock Mos: Towards scaling out-of-distribution detection for large semantic
  space.
\newblock In \emph{Proceedings of the IEEE/CVF Conference on Computer Vision
  and Pattern Recognition}, 8710--8719.

\bibitem[{Krizhevsky et~al.(2009)}]{krizhevsky2009learning}
Krizhevsky, A.; et~al. 2009.
\newblock Learning multiple layers of features from tiny images.

\bibitem[{Kumar et~al.(2022)Kumar, Raghunathan, Jones, Ma, and
  Liang}]{kumar2022fine}
Kumar, A.; Raghunathan, A.; Jones, R.; Ma, T.; and Liang, P. 2022.
\newblock Fine-Tuning can Distort Pretrained Features and Underperform
  Out-of-Distribution.
\newblock In \emph{International Conference on Learning Representations}.

\bibitem[{Li et~al.(2021)Li, Sohn, Yoon, and Pfister}]{li2021cutpaste}
Li, C.-L.; Sohn, K.; Yoon, J.; and Pfister, T. 2021.
\newblock Cutpaste: Self-supervised learning for anomaly detection and
  localization.
\newblock In \emph{Proceedings of the IEEE/CVF Conference on Computer Vision
  and Pattern Recognition}, 9664--9674.

\bibitem[{Liznerski et~al.(2020)Liznerski, Ruff, Vandermeulen, Franks, Kloft,
  and Muller}]{liznerski2020explainable}
Liznerski, P.; Ruff, L.; Vandermeulen, R.~A.; Franks, B.~J.; Kloft, M.; and
  Muller, K.~R. 2020.
\newblock Explainable Deep One-Class Classification.
\newblock In \emph{International Conference on Learning Representations}.

\bibitem[{Melas-Kyriazi et~al.(2022)Melas-Kyriazi, Rupprecht, Laina, and
  Vedaldi}]{melas2022deep}
Melas-Kyriazi, L.; Rupprecht, C.; Laina, I.; and Vedaldi, A. 2022.
\newblock Deep spectral methods: A surprisingly strong baseline for
  unsupervised semantic segmentation and localization.
\newblock In \emph{Proceedings of the IEEE/CVF Conference on Computer Vision
  and Pattern Recognition}, 8364--8375.

\bibitem[{Mirzaei et~al.(2023)Mirzaei, Salehi, Shahabi, Gavves, Snoek,
  Sabokrou, and Rohban}]{mirzaei2023fake}
Mirzaei, H.; Salehi, M.; Shahabi, S.; Gavves, E.; Snoek, C. G.~M.; Sabokrou,
  M.; and Rohban, M.~H. 2023.
\newblock Fake It Until You Make It : Towards Accurate Near-Distribution
  Novelty Detection.
\newblock In \emph{The Eleventh International Conference on Learning
  Representations}.

\bibitem[{Perera and Patel(2019)}]{perera2019learning}
Perera, P.; and Patel, V.~M. 2019.
\newblock Learning deep features for one-class classification.
\newblock \emph{IEEE Transactions on Image Processing}, 28(11): 5450--5463.

\bibitem[{Reiss et~al.(2021)Reiss, Cohen, Bergman, and Hoshen}]{reiss2021panda}
Reiss, T.; Cohen, N.; Bergman, L.; and Hoshen, Y. 2021.
\newblock Panda: Adapting pretrained features for anomaly detection and
  segmentation.
\newblock In \emph{Proceedings of the IEEE/CVF Conference on Computer Vision
  and Pattern Recognition}, 2806--2814.

\bibitem[{Reiss and Hoshen(2023)}]{reiss2023mean}
Reiss, T.; and Hoshen, Y. 2023.
\newblock Mean-shifted contrastive loss for anomaly detection.
\newblock In \emph{Proceedings of the AAAI Conference on Artificial
  Intelligence}, volume~37, 2155--2162.

\bibitem[{Ruff et~al.(2018)Ruff, Vandermeulen, Goernitz, Deecke, Siddiqui,
  Binder, M{\"u}ller, and Kloft}]{ruff2018deep}
Ruff, L.; Vandermeulen, R.; Goernitz, N.; Deecke, L.; Siddiqui, S.~A.; Binder,
  A.; M{\"u}ller, E.; and Kloft, M. 2018.
\newblock Deep one-class classification.
\newblock In \emph{International conference on machine learning}, 4393--4402.
  PMLR.

\bibitem[{Russakovsky et~al.(2015)Russakovsky, Deng, Su, Krause, Satheesh, Ma,
  Huang, Karpathy, Khosla, Bernstein et~al.}]{russakovsky2015imagenet}
Russakovsky, O.; Deng, J.; Su, H.; Krause, J.; Satheesh, S.; Ma, S.; Huang, Z.;
  Karpathy, A.; Khosla, A.; Bernstein, M.; et~al. 2015.
\newblock Imagenet large scale visual recognition challenge.
\newblock \emph{International journal of computer vision}, 115(3): 211--252.

\bibitem[{Sohn et~al.(2020)Sohn, Li, Yoon, Jin, and Pfister}]{sohn2020learning}
Sohn, K.; Li, C.-L.; Yoon, J.; Jin, M.; and Pfister, T. 2020.
\newblock Learning and Evaluating Representations for Deep One-Class
  Classification.
\newblock In \emph{International Conference on Learning Representations}.

\bibitem[{Sun et~al.(2021)Sun, Zhu, Zhang, Zheng, Qiu, Zhang, and
  Wei}]{sun2021dynamic}
Sun, Y.; Zhu, Y.; Zhang, Y.; Zheng, P.; Qiu, X.; Zhang, C.; and Wei, Y. 2021.
\newblock Dynamic Metric Learning: Towards a Scalable Metric Space to
  Accommodate Multiple Semantic Scales.
\newblock In \emph{Proceedings of the IEEE/CVF Conference on Computer Vision
  and Pattern Recognition}, 5393--5402.

\bibitem[{Tack et~al.(2020)Tack, Mo, Jeong, and Shin}]{tack2020csi}
Tack, J.; Mo, S.; Jeong, J.; and Shin, J. 2020.
\newblock CSI: Novelty Detection via Contrastive Learning on Distributionally
  Shifted Instances.
\newblock In \emph{34th Conference on Neural Information Processing Systems
  (NeurIPS) 2020}. Neural Information Processing Systems.

\bibitem[{Tax and Duin(2004)}]{tax2004support}
Tax, D.~M.; and Duin, R.~P. 2004.
\newblock Support vector data description.
\newblock \emph{Machine learning}, 54(1): 45--66.

\bibitem[{Van~der Maaten and Hinton(2008)}]{van2008visualizing}
Van~der Maaten, L.; and Hinton, G. 2008.
\newblock Visualizing data using t-SNE.
\newblock \emph{Journal of machine learning research}, 9(11).

\bibitem[{Van~Gansbeke et~al.(2020)Van~Gansbeke, Vandenhende, Georgoulis,
  Proesmans, and Van~Gool}]{van2020scan}
Van~Gansbeke, W.; Vandenhende, S.; Georgoulis, S.; Proesmans, M.; and Van~Gool,
  L. 2020.
\newblock Scan: Learning to classify images without labels.
\newblock In \emph{European conference on computer vision}, 268--285. Springer.

\bibitem[{Ye, Chen, and Zheng(2021)}]{ye2021understanding}
Ye, Z.; Chen, Y.; and Zheng, H. 2021.
\newblock Understanding the Effect of Bias in Deep Anomaly Detection.
\newblock In \emph{IJCAI 2021: International Joint Conference on Artificial
  Intelligence}.

\bibitem[{Zhang et~al.(2022)Zhang, Yin, Shao, and Liu}]{zhang2022benchmarking}
Zhang, Y.; Yin, Z.; Shao, J.; and Liu, Z. 2022.
\newblock Benchmarking omni-vision representation through the lens of visual
  realms.
\newblock In \emph{European Conference on Computer Vision}, 594--611. Springer.

\end{thebibliography}

\newpage
\clearpage

\appendix

\section{Details for Methods of Comparison}

\begin{enumerate}
	\item PANDA \cite{reiss2021panda} and MSCL \cite{reiss2023mean}: We use the official code and measure performance every 10 epochs. We adjust the learning rate as much as possible to ensure that the performance first improves. The learning rate is between $1e^{-6}$ and the method's original learning rate. We report the best performance in the experiments.
	\item OODWCL \cite{cohen2022out}: We reproduce the method based on the original paper and SCAN \cite{van2020scan}. We select the number of clusters 50 corresponding to the best performance from the preset clustering set $\{10, 50, 100, 200, 500\}$. We also report the best performance during training.
	\item ADIB \cite{deecke2021transfer}: We use the official code. We use the original external dataset, i.e., CIFAR 100 \cite{krizhevsky2009learning}. In the muliti-class experiments, when the class num is less than 32, we measure performance every 10 epochs. When the class num is greater than 32, we measure performance every 1 epoch. At the same time, we adjust the learning rate to $1e^{-6}$. We report the best performance in the experiments.
	\item FITYMI \cite{mirzaei2023fake}: We report the results in the original paper.
\end{enumerate}

All experiments were conducted on Ubuntu 18.04.5 and a computer equipped with Xeon(R) Gold 6140R CPUs@2.30GHZ and 8 NVIDIA GeForce RTX 2080 Ti with 11 GB of memory.

\renewcommand{\thetable}{8}
\begin{table}[h]
    \centering
    \caption{Details of different experimental datasets. $c_{train}$ denotes the number of categories of data in the training set. $c_{test}$ denotes the number of categories of data in the test set, including one-class samples and outliers (The number of categories in front of $+$ indicates the number of categories of one-class samples and the number of categories of outliers in the back). $N_{train}$ denotes the number of one-class samples in the training set. $N_{test}$ denotes the number of samples in the test set. ND denotes the near distribution experiment. MC 2 denotes multi-class experiments with 2 categories.}
    \begin{tabular}{c c c c c}
    \hline
    Dataset & $c_{train}$ & $c_{test}$ & $N_{train}$ & $N_{test}$ \\ \hline
    \multicolumn{3}{l}{\shortstack{Single Class Experiments}} & \\
    CIFAR 10 & 1 & 1+9 & 5000 & 10000 \\
    ND & 1 & 1+1 & 10031 & 29759 \\ \hline
    \multicolumn{3}{l}{\shortstack{Multi-class Experiments}} & \\
    MC 2 & 2 & 2+64 & 694 & 3899 \\ 
    MC 4 & 4 & 4+64 & 1331 & 3939 \\
    MC 8 & 8 & 8+64 & 2701 & 4019 \\
    MC 16 & 16 & 16+64 & 4970 & 4178 \\ 
    MC 32 & 32 & 32+64 & 8898 & 4497 \\
    MC 64 & 64 & 64+64 & 15823 & 5137 \\ 
    MC 128 & 128 & 128+64 & 27646 & 6413 \\ 
    MC 256 & 256 & 256+64 & 46708 & 8957 \\ 
    MC 512 & 512 & 512+64 & 76198 & 14048 \\ 
    MC 1024 & 1024 & 1024+64 & 115692 & 23544 \\ \hline
    \end{tabular}
    \label{Appendix_2}
\end{table}

\renewcommand{\thetable}{9}
\begin{table*}[h]
\footnotesize
    \centering
    \caption{Per-class one-class classification performance (AUROC \%) on CIFAR 10. * denotes that the features are $\rm{L}_{2}$ normalized.}
    \begin{tabular}{c c c c c c c c c c c c}
    \hline
    & 0 & 1 & 2 & 3 & 4 & 5 & 6 & 7 & 8 & 9 & mean \\ \hline
    ${\rm CA}^{2}$ & 96.9 & 98.5 & 93.2 & 91.1 & 96.7 & 95.4 & 96.7 & 97.5 & 97.8 & 97.7 & 96.1 \\
    ${\rm CA}^{2}$* & 97.5 & 98.6 & 95.0 & 94.1 & 96.9 & 97.2 & 97.7 & 98.2 & 98.3 & 98.4 & 97.2 \\ \hline
    \end{tabular}
    \label{Appendix_1}
\end{table*}


\section{Details for Multi-class Experiments}

We use the dataset in OmniBenchmark \cite{zhang2022benchmarking}. It includes 21 realm-wise datasets with 7,372 concepts and 1,074,346 images. To ensure that the semantics of the training data do not overlap with the semantics of the outliers, we extract the data in 7 realms, including mammal, instrumentality, device, consumer goods, car, structure and food. There is a similar semantic space between the used samples and ImageNet-1K. We evenly select classes from 7 realms according to the number of each category from more to less. The outlier is the NINCO (No ImageNet Class Objects) dataset \cite{bitterwolf2023or}, where there are no ImageNet-1K class objects. It consists of 64 out of distribution of ImageNet-1K classes. More details are listed in Table \ref{Appendix_2}.

\section{Details for Pre-trained Models in Ablation}

For the supervised pre-training, we use a pre-trained ViT Base model \cite{dosovitskiy2020image}. For self-supervised pre-trianing, we use a ViT-Base model and a ResNet 50 model pre-trained by MoCo v3 \cite{chen2021empirical}.

\section{Supplementary Experimental Results}

The per-class performance of CIFAR 10 are listed in Table \ref{Appendix_1}.

\end{document}